\documentclass{article}

\usepackage{arxiv}

\usepackage[utf8]{inputenc} 
\usepackage[T1]{fontenc}    
\usepackage{hyperref}       
\usepackage{url}            
\usepackage{booktabs}       
\usepackage{amsfonts}       
\usepackage{nicefrac}       
\usepackage{microtype}      
\usepackage{graphicx}
\usepackage{natbib}
\usepackage{doi}
\usepackage[table]{xcolor}
\usepackage{colortbl}
\usepackage{setspace}
\usepackage{tikz}
\usepackage{adjustbox}
\usepackage{caption}
\usetikzlibrary{shadows,shapes,arrows,positioning,fit,shapes.geometric}
\usepackage{varwidth} 

\usepackage{amssymb}
\usepackage{amsmath}

\definecolor{PriceShareColor}{RGB}{173,216,230}   
\definecolor{EarthquakeColor}{RGB}{144,238,144}   
\definecolor{UniMiBColor}{RGB}{255,255,204}       
\definecolor{DJIColor}{RGB}{255,204,153}          
\definecolor{MillColor}{RGB}{204,153,255}         
\definecolor{ECG5000Color}{RGB}{153,255,255}      

\title{Time Series Embedding Methods for Classification Tasks: A Review} 

\author{
\href{https://orcid.org/0009-0003-5035-9257}{\includegraphics[scale=0.06]{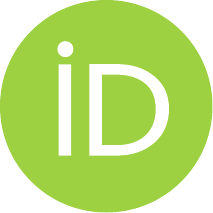}\hspace{1mm}Habib Irani}\qquad
Yasamin Ghahremani \qquad
Arshia Kermani \qquad
\href{https://orcid.org/0000-0002-7371-8887}{\includegraphics[scale=0.06]{orcid.pdf}\hspace{1mm}Vangelis Metsis} \\
Department of Computer Science \\
Texas State University \\
San Marcos, TX 78666 \\
\texttt{\{habibirani, zub11, arshia.kermani, vmetsis\}@txstate.edu}
}

\begin{document}
\maketitle

\begin{abstract}
Time series analysis has become crucial in various fields, from engineering and finance to healthcare and social sciences. Due to their multidimensional nature, time series often need to be embedded into a fixed-dimensional feature space to enable processing with various machine learning algorithms. In this paper, we present a comprehensive review and quantitative evaluation of time series embedding methods for effective representations in machine learning and deep learning models. We introduce a taxonomy of embedding techniques, categorizing them based on their theoretical foundations and application contexts. Our work provides a quantitative evaluation of representative methods from each category by assessing their performance on downstream classification tasks across diverse real-world datasets. Our experimental results demonstrate that the performance of embedding methods varies significantly depending on the dataset and classification algorithm used, highlighting the importance of careful model selection and extensive experimentation for specific applications. To facilitate further research and practical applications, we provide an open-source code repository\footnote{Source code: \url{https://github.com/imics-lab/time-series-embedding}} implementing these embedding methods. This study contributes to the field by offering a systematic comparison of time series embedding techniques, guiding practitioners in selecting appropriate methods for their specific applications, and providing a foundation for future advancements in time series analysis.
\end{abstract}


\section{Introduction}
Time series embedding is a technique used to represent time series data in the form of vector embeddings. Today, time series analysis methods have emerged as a fundamental element across a vast amount of applications ranging from finance, as in the work of \cite{Zhu2022}, to healthcare, as demonstrated in \cite{nejedly2022classification,MORID2021,Chen2021,Lee2021,Soenksen2022}, engineering applications such as machine health monitoring, predictive maintenance, and fault detection \cite{zhao2019deep,li2020systematic}, and social sciences, explored by \cite{SANTOSH2018}. As machine learning and deep learning techniques continue to advance, there is a growing need for effective methods to represent and analyze time series data in these models. Tasks such as anomaly detection, classification, pattern recognition, prediction, and decision-making now heavily rely on robust methods that could accurately embed these often high-dimensional data into scalable yet informative representations.

The importance of studying and evaluating different time series embedding methods stems from several key factors:

\begin{itemize}
\item \emph{Dimensionality reduction}: Time series data often has high dimensionality, which can lead to computational challenges and the curse of dimensionality. Effective embedding methods can reduce the dimensionality while preserving essential temporal patterns and relationships.
\item \emph{Feature extraction}: Embeddings can automatically extract relevant features from raw time series data, potentially capturing complex temporal dependencies that may not be apparent in the original representation.
\item \emph{Improved model performance}: Well-designed embeddings can lead to significant improvements in the performance of downstream machine learning tasks, such as classification, clustering, and forecasting.
\item \emph{Transfer learning}: Embeddings learned from large datasets can be transferred to smaller, related datasets, enabling more effective learning in scenarios with limited data.
\item \emph{Interpretability}: Some embedding methods can provide insights into the underlying structure and patterns of time series data, aiding in data exploration and understanding.
\item \emph{Handling irregularities}: Many real-world time series datasets are characterized by irregular sampling, missing values, or varying lengths. Certain embedding methods can address these challenges more effectively than others.
\end{itemize}

As the field of time series analysis continues to evolve, a wide array of embedding methods has been proposed, each with its own strengths and limitations. These methods range from classical approaches like delay embeddings and Fourier transforms to more recent techniques leveraging deep learning architectures such as recurrent neural networks and transformer models.
Given the diversity of available methods and their potential impact on downstream applications, a comprehensive evaluation and comparison of time series embedding techniques is crucial. This survey aims to provide an overview of the current landscape of time series embedding methods, assess their representation strength when combined with various classification algorithms, and offer insights into selecting appropriate embedding techniques for specific applications.

\begin{figure}
\centering
\includegraphics[width=\textwidth]{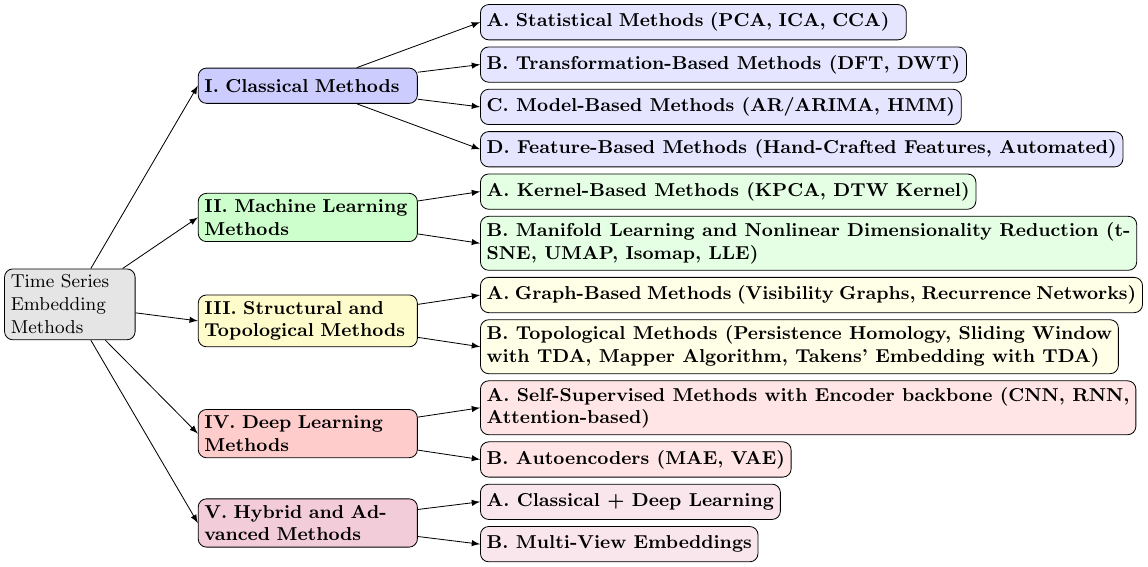}
\caption{Taxonomy of Time Series Embedding Methods}
\label{fig:taxonomy}
\end{figure}

Creating a taxonomy for time series embedding methods can be approached in several different ways, depending on the criteria or perspectives one chooses to emphasize. Those can be based on the theoretical foundations or mathematical principles used, domain of information captured, model complexity and computational requirements, scalability and data requirements, nature of time series data (uni-/muti-variate), application context, etc. In this work, we choose to categorize embeddings mainly based on their theoretical foundations and application context, creating a taxonomy of different categories as depicted in Figure~\ref{fig:taxonomy} and in more detail in Table~\ref{tab:taxonomy}.

\begingroup
\renewcommand{\arraystretch}{1.5}
\begin{table}[htbp]
\centering
\caption{Detailed Categories of Time Series Embedding Methods}
\label{tab:taxonomy}

\begin{small}
\begin{tabular}{|p{3cm}|p{11.5cm}|}
\hline
\textbf{Category} & \textbf{Representative Examples} \\ \hline

\textbf{Statistical} & 
\textbf{PCA:} Reduces dimensionality by identifying orthogonal axes with maximum variance. \textbf{ICA:} Decomposes the series into statistically independent components. \textbf{CCA:} Identifies linear relationships between two sets of variables, revealing common patterns. \\ \hline

\textbf{Transformation-Based} & 
\textbf{DFT:} Transforms the series into frequency components, using dominant frequencies as embeddings. \textbf{DWT:} Captures time and frequency characteristics using wavelet coefficients. \\ \hline

\textbf{Feature-Based} & 
\textbf{Hand-Crafted:} \textit{Statistical:} Extracts mean, variance, skewness, etc. \textit{Time-Domain:} Identifies peaks, troughs, zero-crossings. \textit{Frequency-Domain:} Captures spectral power, dominant freqs. \\
& \textbf{Automated:} \textit{TSFRESH:} Extracts a wide range of features automatically. \textit{catch22:} Provides 22 efficient time series characteristics. \\ \hline

\textbf{Model-Based} & 
\textbf{AR/ARMA/ARIMA:} Uses past values and moving averages to model the series. \textbf{HMM:} Represents the series as a sequence of hidden states with probabilistic transitions. \\ \hline

\textbf{Kernel-Based} & 
\textbf{KPCA:} Extends PCA with kernel methods for non-linear relationships. \textbf{DTW Kernel:} Measures similarity between series, accounting for temporal distortions. \\ \hline

\textbf{Graph-Based} & 
\textbf{Visibility Graphs:} Converts data into a graph, with embeddings from graph properties. \textbf{Recurrence Networks:} Uses recurrence plots to construct networks for embedding. \\ \hline

\textbf{Manifold Learning and Nonlinear Dimensionality Reduction} & 
\textbf{t-SNE:} Preserves local structure in lower-dimensional embeddings. \textbf{UMAP:} Provides non-linear embeddings while preserving structure. \textbf{Isomap:} Captures intrinsic geometry by preserving geodesic distances. \textbf{LLE:} Maps the series onto a lower-dimensional manifold, preserving local structure. \\ \hline

\textbf{Topological} & 
\textbf{Persistence Homology:} Captures topological features across scales using persistence diagrams. \textbf{Sliding Window with TDA:} Applies TDA on time-delay embeddings to capture dynamics.  \textbf{Mapper Algorithm:} Constructs a topological network representing the data's shape. \textbf{Takens' Embedding with TDA:} Reconstructs the phase space and applies TDA. \\ \hline

\textbf{Deep Learning-Based} & 
\textbf{Autoencoders:} Compress and reconstruct series, with embeddings from the bottleneck layer. \textbf{RNNs:} Capture temporal dependencies using hidden state embeddings. \textbf{CNNs:} Extract local patterns through convolution, creating feature embeddings. \textbf{Attention-Based Models:} Focus on relevant parts of the series for embedding. \\ \hline

\textbf{Hybrid} & 
\textbf{Classical + Deep Learning:} Combines traditional methods with deep learning for robust embeddings. \textbf{Multi-View Embeddings:} Integrates multiple perspectives, transformations, or models. \\ \hline

\end{tabular}
\end{small}
\end{table}
\endgroup

Previous surveys on time series embeddings, such as the one published by \cite{tjostheim2023some}, have provided a qualitative categorization of the various methods but have not quantitatively evaluated the representation capability of each method on real-world data. In this work, we evaluate popular time series embedding methods by using the formed embeddings on downstream classification tasks, which provides a crucial perspective on their effectiveness and generalization capabilities. Classification tasks serve as an excellent proxy for assessing how well embeddings capture discriminative features and preserve relevant temporal patterns. Unlike forecasting, which focuses primarily on predictive accuracy, or clustering and anomaly detection, which rely heavily on unsupervised learning, classification offers a supervised framework that allows for a more direct and interpretable evaluation of embedding quality. By using labeled data, we can quantitatively measure how well the embeddings separate different classes of time series, which is often a key requirement in real-world applications with both traditional (KNN, SVM, Random Forest, Gradient Boosting, etc.) and deep learning-based methods. Furthermore, classification tasks typically have well-established evaluation metrics and benchmarks, facilitating comparisons across different embedding methods. This approach also aligns well with the common use case of using pre-trained embeddings as input features for various downstream tasks, where classification is frequently encountered. 

Our experimental evaluation shows that the representation capabilities of various embedding methods can vary across different datasets and classification algorithms. This emphasizes the need for extensive experimentation and model selection to highlight the best combination of embedding and classification algorithms for the particular task at hand. Along with this evaluation, we provide an open-source suite
that implements these embedding methods for use by the research community. 

The remainder of this paper is organized as follows. In section~\ref{sec:background}, we provide a brief overview of the different time series embedding categories that form our taxonomy as a background. In section~\ref{sec:methodology}, we detail the machine learning pipeline that we followed to evaluate each method quantitatively as well as a more detailed theoretical description of each embedding method evaluated in this study. In section~\ref{sec:results}, we present the experimental results along with a discussion of our observations. Finally, section~\ref{sec:conclusion} concludes this paper.

\section{Background}
\label{sec:background}

Time series embedding methods have evolved significantly, driven by the need to represent complex temporal data in a form suitable for various machine learning tasks. These methods can be broadly categorized into the following main groups: Statistical, Transformation-Based, Feature-Based, Model-Based, Kernel-Based, Graph-Based, Manifold Learning and Nonlinear Dimensionality Reduction, Topological, Deep Learning-Based, and Hybrid methods. Each category represents a distinct approach to embedding, with unique strengths and weaknesses. 

\subsection{Statistical Methods}
Statistical methods have been fundamental to time series analysis for decades. Principal Component Analysis (PCA), as established in the foundational works of \cite{pearson1901liii,hotelling1933analysis}, is one of the earliest techniques that reduces dimensionality by identifying orthogonal axes with maximum variance, allowing for a compact representation of time series data. Building on this work, research by \cite{comon1994independent} introduced Independent Component Analysis (ICA), which extends this by decomposing time series into statistically independent components, particularly useful in fields like neuroscience and signal processing, where uncovering hidden sources is essential. The work of \cite{hotelling1992relations} developed Canonical Correlation Analysis (CCA), which identifies linear relationships between two sets of variables, making it valuable for capturing common patterns across multiple time series. As demonstrated in the work of \cite{klein1997statistical}, these methods provide robust, interpretable embeddings that serve as a strong foundation for more complex analyses or as standalone tools for time series exploration.

\subsection{Transformation-Based Methods}
Transformation-based methods like the Fourier Transform (FT) and Wave-let Transform (WT) have been instrumental in revealing patterns within time series data that are not visible in the time domain alone, as shown in the work of \cite{michau2022fully}. According to the analysis of \cite{sneddon1995fourier}, the Fourier Transform decomposes a series into its constituent frequencies, making it suitable for analyzing periodic components. However, it assumes stationarity, limiting its effectiveness for non-stationary data. The seminal works of \cite{morlet1982wave,grossman1985decomposition,meyer1993wavelets} introduced the Wavelet Transform as a more versatile alternative, capturing both time and frequency information, making it more suitable for analyzing non-stationary and transient signals.

\subsection{Feature-Based Methods}
Feature-based methods involve extracting key characteristics from time series data, either manually or automatically. Hand-crafted features can include statistical measures like mean and variance, or more complex time-domain and frequency-domain features. Recent advances such as TSFRESH by \cite{christ2018time} and catch22 by \cite{lubba2019catch22} provide a more systematic approach to feature extraction, offering a wide range of features tailored to different types of time series data \cite{christ2018time,lubba2019catch22}. These methods are particularly useful in scenarios where domain knowledge is limited, allowing for the extraction of informative features without manual intervention.

\subsection{Model-Based Methods}
Model-based methods represent time series as sequences of states or as outputs of generative models. As explored in the works of \cite{BUXTON2019,harvey1990arima}, Autoregressive (AR) and ARIMA models are traditional examples, while more complex methods like Hidden Markov Models (HMMs) capture the probabilistic transitions between different states in the series. Even though autoregressive methods are often classified as statistical processes, due to the fact that they are built on statistical concepts like autocorrelation and moving averages, these methods explicitly model the underlying process generating the time series, assuming a specific structure for the data-generating process and creating a mathematical model of the time series for forecasting and analysis. These models are powerful for time series with underlying state-based dynamics but require assumptions about the underlying processes, which may not always hold.

\subsection{Kernel-Based Methods}
Kernel-based methods extend classical statistical techniques like PCA to capture non-linear relationships within time series data. The foundational work of \cite{scholkopf1997kernel} introduced Kernel PCA, which projects data into a higher-dimensional space where linear separation becomes possible. Building on this approach, research by \cite{berndt1994using} developed techniques like the Dynamic Time Warping (DTW) kernel to measure similarity between time series by accounting for temporal distortions, making them robust to variations in speed and amplitude. These methods are effective in capturing complex, non-linear structures in the data but can be computationally intensive.

\subsection{Graph-Based Methods}
Graph-based methods, including Visibility Graphs and Recurrence Networks, convert time series data into graphical representations where the nodes represent data points, and edges represent relationships between them. As demonstrated by \cite{lacasa2008time}, these methods leverage graph theory to analyze the structural properties of time series, offering insights that traditional methods may overlook. According to the work of \cite{donner2010recurrence,lacasa2008time}, visibility graphs transform a time series into a graph by connecting nodes based on their visibility, while recurrence networks analyze the recurrence of states within the series. Recent work by \cite{kutluana2024classification} has used these concepts for studying complex time series data, discussing how methods such as visibility graphs appear to be robust to noise. As shown by \cite{liu2015differ}, contrary to other embedding methods, the visibility graph formation does not require the tuning of its parameters. These methods are also particularly useful in studying the underlying dynamics of complex systems.

\subsection{Manifold Learning and Nonlinear Dimensionality Reduction}
Manifold learning methods, as developed by \cite{roweis2000nonlinear}, \cite{van2008visualizing}, \cite{tenenbaum2000global}, and \cite{mcinnes2018umap}, including approaches like Locally Linear Embedding (LLE), t-SNE, Isomap, and UMAP, are designed to uncover the underlying structure of high-dimensional time series data by preserving local and global geometric properties in a lower-dimensional space \cite{roweis2000nonlinear,van2008visualizing,tenenbaum2000global,mcinnes2018umap}. These methods are particularly effective for visualizing high-dimensional data and for capturing complex, non-linear relationships that traditional linear methods cannot handle. However, they may require careful tuning of parameters and are sensitive to noise and uneven sampling.

\subsection{Topological Methods}
Topological Data Analysis (TDA) offers a unique perspective by capturing the shape of data. As explored in the works of \cite{edelsbrunner2002topological,singh2007topological}, techniques like Persistent Homology and the Mapper Algorithm focus on identifying topological features that are stable across different scales of analysis. These methods are valuable for understanding the global structure of time series data, particularly in applications where the shape of data plays a crucial role, such as in dynamical systems and complex networks.

\subsection{Deep Learning-Based Methods}
Deep learning methods have revolutionized time series embedding by leveraging neural networks to learn complex, hierarchical representations. The work of \cite{hochreiter1997long} introduced Recurrent Neural Networks (RNNs) and their variants like Long Short-Term Memory (LSTM) networks, which are particularly suited for capturing temporal dependencies. As shown by \cite{krizhevsky2017imagenet}, Convolutional Neural Networks (CNNs), originally designed for image processing, have also been adapted for time series by treating the series as a one-dimensional grid. More recently, research by \cite{vaswani2017attention} demonstrated how attention-based models like Transformers show promise in modeling long-range dependencies in time series data. These methods excel in tasks where large amounts of labeled data are available but may suffer from overfitting and require significant computational resources.

\subsection{Hybrid Methods}
Hybrid methods combine the strengths of multiple embedding techniques to address the limitations of individual methods. As demonstrated by \cite{Li2022}, combining statistical methods with deep learning can enhance interpretability while retaining the powerful feature extraction capabilities of neural networks. Other approaches integrate multiple perspectives, such as combining time-domain and frequency-domain features, or using graph-based embeddings alongside traditional machine learning models. Hybrid methods are often tailored to specific applications, making them versatile but potentially complex to implement.

The diverse landscape of time series embedding methods offers a rich toolkit for researchers and practitioners. Each category of methods has its strengths and limitations, making the choice of embedding technique highly dependent on the specific characteristics of the data and the requirements of the downstream task. As the field continues to evolve, new methods and hybrid approaches are likely to emerge, further expanding our ability to extract meaningful representations from time series data.


\section{Evaluation Methodology}
\label{sec:methodology}

In this section, we detail the methodology used to evaluate the effectiveness of various time series embedding methods. Our approach involves systematically comparing the most popular of these methods across different datasets and classification tasks to assess their ability to capture and represent the essential characteristics of temporal data. The evaluation is conducted through a machine learning pipeline, encompassing data preprocessing, embedding generation, and subsequent model training and validation. The following subsections detail each component of our evaluation process, including the datasets utilized, and the machine learning pipeline implemented to assess classification performance and theoretical definition of the specific embedding methods examined.

\subsection{Data}

This paper explores a variety of time series with different characteristics. Table~\ref{tab:datasets} presents the properties of the datasets used to evaluate the embedding methods discussed in this research. Data was sourced from various open repositories, such as the Time Series Classification Repository \cite{TSclassi} and the UC Irvine Machine Learning Repository \cite{UCI}.

\begin{table}
\centering
\caption{Properties of time series datasets used in this study.}
\label{tab:datasets}

\begin{adjustbox}{width=0.8\columnwidth}
\centering
\setlength{\tabcolsep}{4pt} 
\begin{tabular}{@{}lcccccc@{}}
\toprule
\textbf{Dataset} & \textbf{Train Size} & \textbf{Test Size} & \textbf{Length} & \textbf{Classes} & \textbf{Channels} & \textbf{Type} \\
\midrule
Sleep & 478,785 & 90,315 & 178 & 5 & 1 & EEG \\
ElectricDevices & 8,926 & 7,711 & 96 & 7 & 1 & Device \\
MelbournePedestrian & 1,194 & 2,439 & 24 & 10 & 1 & Traffic \\
RacketSports & 151 & 152 & 30 & 4 & 6 & HAR \\
SharePriceIncrease & 965 & 965 & 60 & 2 & 1 & Financial \\
SelfRegulationSCP1 & 268 & 293 & 896 & 2 & 6 & EEG \\
UniMiB-SHAR & 4,601 & 1,524 & 151 & 9 & 3 & HAR \\
EMGGestures & 1,800 & 450 & 30 & 8 & 9 & EMG \\
Mill &7751&  1910 &64 & 3 & 6 & Sensor \\
ECG5000 & 500 & 4500 & 140 & 5 & 1 & ECG \\
\bottomrule
\end{tabular}
\end{adjustbox}
\end{table}

\begin{enumerate}
  \item \emph{Sleep:} Originally from PhysioNet’s “Sleep EDF” database, this dataset comprises 153 whole-night single-lead EEG recordings (100 Hz) from 82 healthy subjects. Recordings are segmented into non-overlapping 178-sample epochs labeled as five sleep stages (Wake, N1, N2, N3, REM). We use the split of 478,785 train+validation and 90,315 test, noting class imbalance across partitions.
  \item \emph{ElectricDevices:} Drawn from the UCR Time Series Classification Archive, these series capture appliance power consumption sampled every two minutes over one month in 251 UK households. Each of the 8,926 training and 7,711 test series is 96 samples long and classified into seven usage profiles.
  \item \emph{MelbournePedestrian:} From the City of Melbourne’s automated pedestrian counting system, this dataset contains 24 hourly counts per day at ten locations during 2017. We treat each 24-sample day as one series, using 1,194 days for training and 2,439 for testing, with class labels corresponding to sensor sites.
  \item \emph{RacketSports:} Recorded at 10 Hz via a wrist-worn Sony SmartWatch 3, each 30-sample series encodes accelerometer (x,y,z) then gyroscope (x,y,z) readings over a 3 s racket stroke. There are 151 train and 152 test instances labeled as one of four actions (badminton clear/smash, squash forehand/backhand).
  \item \emph{SharePriceIncrease:} Formatted from daily NASDAQ-100 closing prices, each 60-day series records percentage change from the prior day. The binary label indicates whether the stock rose more than 5\% after its next quarterly earnings release (0 = no, 1 = yes). We have 965 train and 965 test series.
  \item \emph{SelfRegulationSCP1:} A slow cortical potentials BCI dataset recorded at 256 Hz over six EEG channels during a cursor-control task. Each 896-sample trial (3.5 s feedback window) comprises 268 train and 293 test trials, labeled by intentional cortical positivity vs. negativity.
  \item \emph{UniMiB-SHAR:} Collected via a smartphone accelerometer at 50 Hz, this dataset contains 4,601 training and 1,524 testing tri-axial series, each 151 samples long, capturing nine daily activities and falls from 30 subjects (ages 18–60).
  \item \emph{EMGGestures:} Recorded by a nine-channel EMG armband at 50 Hz, this dataset comprises 1,800 train and 450 test signals of length 30, classified into eight hand gestures (e.g., fist, wave-in, pinch).
  \item \emph{Mill:} From NASA’s milling-machine sensor suite, each 64-sample series contains readings from six sensors (acoustic emission, vibration, current) under varying cutting conditions. There are 7 751 train and 1,910 test instances across three tool-wear classes.
  \item \emph{ECG5000:} A pre-processed subset of PhysioNet’s BIDMC CHF Database (“chf07”), where individual heartbeats were extracted from a 20 h ECG and interpolated to 140 samples. We select 500 train and 4,500 test beats, labeled into five heartbeat-type classes.
\end{enumerate}

These diverse datasets allow us to evaluate the performance of our embedding methods across different domains and time series characteristics.

\subsection{Machine Learning Pipeline}

\begin{figure}
\centering
\includegraphics[width=\textwidth]{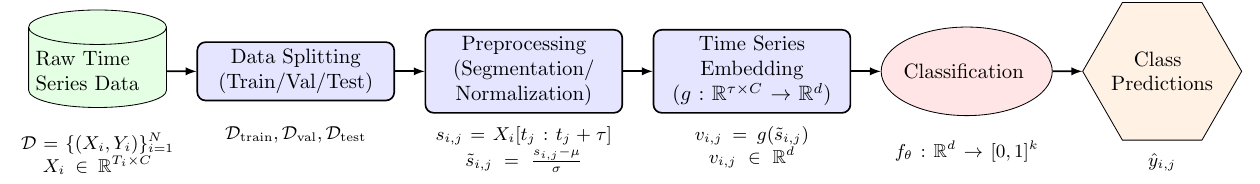}
\caption{Machine learning pipeline for time series classification.}
\label{fig:ml_pipeline}
\end{figure}

We consider a dataset  \(\mathcal{D} = \{(X_i, Y_i)\}_{i=1}^N\), where each \(X_i \in \mathbb{R}^{T_i \times C}\) is a multi-channel, continuous time series with \(T_i\) time steps and \(C\) channels. Associated with each time series \(X_i\) is a sequence of labels \(Y_i \in \mathcal{L}^{T_i}\), with \(\mathcal{L}\) representing the set of possible labels. The dataset is suitable for supervised learning tasks involving time series classification, applicable to diverse scenarios such as physiological data, air quality monitoring, and activity recognition using wearable devices. The machine learning pipeline we follow to evaluate our embedding methods is summarized in Figure~\ref{fig:ml_pipeline} and described in detail in the following subsections.

\subsubsection{Data Splitting:}

Before processing, the dataset \(\mathcal{D}\) is divided into training (\(\mathcal{D}_{\textnormal{train}}\)), validation (\(\mathcal{D}_{\textnormal{val}}\)), and test (\(\mathcal{D}_{\textnormal{test}}\)) subsets. This split is performed to ensure that the data from a single entity (e.g., a specific subject or period) is exclusively contained within one of these subsets, maintaining complete independence between the training, validation, and test sets. 

\subsubsection{Time Series Segmentation:}

After the dataset is split, each subset (\(\mathcal{D}_{\textnormal{train}}\), \(\mathcal{D}_{\textnormal{val}}\), \(\mathcal{D}_{\textnormal{test}}\)) undergoes a segmentation process. Let \(\tau\) be the window size and \(\omega\) the overlap between consecutive windows, both defined as hyperparameters. For each time series \(X_i\) in a subset, we segment it into windows:
\[
s_{i,j} = X_i[t_j : t_j + \tau], \quad t_j = 1, \tau - \omega + 1, 2(\tau - \omega) + 1, \ldots, T_i - \tau + 1
\]
The corresponding labels for each segment \(s_{i,j}\) are determined by an aggregation function applied to \(Y_i\) over the window:
\[
y_{i,j} = \textnormal{aggregation}(Y_i[t_j : t_j + \tau])
\]

In this work, the label aggregation function used was based on the mode of the label of the data in that segment.

\subsubsection{Data Normalization:}

Each segment \(s_{i,j}\) from \(\mathcal{D}_{\textnormal{train}}\), \(\mathcal{D}_{\textnormal{val}}\), and \(\mathcal{D}_{\textnormal{test}}\) is preprocessed through a normalization function \(f\). The normalized segment is denoted as \(\tilde{s}_{i,j}\). That normalization is commonly a standardization to zero mean and unit variance:

\[
\tilde{s}_{i,j}[t, c] = f(s_{i,j}[t, c]) = \frac{s_{i,j}[t, c] - \mu_c}{\sigma_c}
\]
where \(\mu_c\) and \(\sigma_c\) are the mean and standard deviation of channel (or feature) \(c\) computed over the training segments.

Alternatively, a min-max scaling can be applied. This is calculated through:
\[
\tilde{s}_{i,j}[t, c] = f(s_{i,j}[t, c]) = \frac{s_{i,j}[t, c] - \min_c}{\max_c - \min_c}
\]
where \(\min_c\) and \(\max_c\) are the minimum and maximum values of channel \(c\) in the training set.

\subsubsection{Time Series Embedding:}
After preprocessing, each segment \(\tilde{s}_{i,j}\) is transformed into an embedding vector \(v_{i,j}\) using a predefined embedding function \(g\):
\[
v_{i,j} = g(\tilde{s}_{i,j}), \quad g: \mathbb{R}^{\tau \times C} \to \mathbb{R}^d
\]

Each embedding vector \(v_{i,j} \in \mathbb{R}^d\) is then used as an input instance to the machine learning classification algorithm.

\subsubsection{Model Training, Validation, and Testing}

The embedded vectors \(\{v_{i,j}\}\) from each subset are used to train, validate, and test a machine learning model. The training set \(\mathcal{D}_{\textnormal{train}}\) is used for model learning, while the validation set \(\mathcal{D}_{\textnormal{val}}\) assists in hyperparameter tuning. The model's performance is subsequently evaluated using the embedded test set \(\mathcal{D}_{\textnormal{test}}\), with outcomes measured by metrics such as classification accuracy. 

We have applied classical and neural network-based time series classification methods to explore the classification results. In particular, Logistic Regression, Decision Trees, Random Forest, K-Nearest Neighbors (KNN), XGBoost, Support Vector Machines (SVM), Naive Bayes, and Multi-Layer Perceptron (MLP) classification methods have been used to study the performance and accuracy of the embedding methods discussed in the paper.

\subsection{Embedding Methods Evaluated}

In this subsection, we examine in more detail the embedding methods that were selected for comprehensive evaluation. These methods were selected based on their popularity while representing as many of the different categories from our taxonomy as possible. To keep the embedding process independent of the downstream classification task, we opted for using only unsupervised techniques for creating the embeddings, i.e. no labels were used during the mapping of the raw time series data into an embedding vector. Labels were used only when training the final classifier on the previously created embedding vectors.

\subsubsection{Principal Component Analysis (PCA):}

PCA is a technique that transforms a set of correlated variables into a smaller set of uncorrelated variables called principal components. The first principal component captures the most variance in the data, the second principal component captures the second most variance, and so on. The formula for PCA is: \(X = U \Sigma V^\top\), where \(X\) is the input data matrix, \(U\) contains the left singular vectors, \(\Sigma\) is a diagonal matrix of singular values, and \(V^\top\) contains the right singular vectors.

The embedding process with PCA operates as follows:

\begin{enumerate}
    \item The normalized segments \(\tilde{s}_{i,j}\) are vectorized (flattened) into one-dimensional vectors:
    \(
    \mathbf{\tilde{s}}_{i,j} = \text{vec}(\tilde{s}_{i,j}) \in \mathbb{R}^{\tau C}
    \)
    \item These vectors are organized into a data matrix \(X \in \mathbb{R}^{n_s \times \tau C}\), where each row corresponds to a vectorized segment, and \(n_s\) is the total number of segments in the training set.
    \item PCA is applied to \(X\) to obtain the projection matrix \(W \in \mathbb{R}^{\tau C \times d}\), whose columns are the top \(d\) principal components.
    \item Each segment is transformed into its embedding using the PCA embedding function:
    \(
    v_{i,j} = g(\tilde{s}_{i,j}) = W^\top \mathbf{\tilde{s}}_{i,j}
    \)
    \item The resulting vectors \(v_{i,j} \in \mathbb{R}^d\) serve as the embedded representations of the original time series segments.
\end{enumerate}

Other embedding methods follow a similar process, and the embedding steps will be omitted for brevity.

\subsubsection{Fourier Transform (FFT):}

The Fourier Transform decomposes a time series into its constituent frequencies. For each normalized segment \( \tilde{s}_{i,j} \), we apply the Discrete Fourier Transform (DFT) to obtain its frequency representation. For a univariate time series \( x_n \), the DFT and its inverse are given by:

\[
X_k = \sum_{n=0}^{N-1} x_n e^{-i 2 \pi k n / N}, x_n = \frac{1}{N} \sum_{k=0}^{N-1} X_k e^{i 2 \pi k n / N}
\]

where \( x_n \) is the input signal at time step \( n \), \( X_k \) is the DFT coefficient at frequency \( k \), \( N \) is the length of the signal, and \( i \) is the imaginary unit.

For multivariate time series segments \( \tilde{s}_{i,j} \in \mathbb{R}^{\tau \times C} \), we apply the DFT independently to each channel \( c \) to obtain the frequency components \( X_{i,j}^{(c)} \). The embedding vector \( v_{i,j} \) is then formed by concatenating the magnitudes (or other features) of the DFT coefficients from each channel:

\[
v_{i,j} = g(\tilde{s}_{i,j}) = \text{concat}\left( |X_{i,j}^{(1)}|, |X_{i,j}^{(2)}|, \ldots, |X_{i,j}^{(C)}| \right)
\]

where \( g \) is the embedding function, and \( |X_{i,j}^{(c)}| \) denotes the magnitude spectrum of channel \( c \).

\subsubsection{Wavelet Transform:}

The Wavelet Transform decomposes a time series into time-frequency representations at different scales. For each normalized segment \( \tilde{s}_{i,j} \), we apply the Continuous Wavelet Transform (CWT) to capture both time and frequency information. The CWT of a signal \( x(t) \) is defined as:

\[
CWT(a, b) = \frac{1}{\sqrt{|a|}} \int_{-\infty}^{\infty} x(t) \psi^*\left(\frac{t - b}{a}\right) dt
\]

where \( \psi(t) \) is the mother wavelet, \( a \) is the scale parameter, \( b \) is the translation parameter, and \( \psi^* \) denotes the complex conjugate of \( \psi \).

For multivariate segments \( \tilde{s}_{i,j} \), the CWT is applied independently to each channel \( c \). The embedding \( v_{i,j} \) is constructed by extracting features from the wavelet coefficients, such as energies at different scales or statistical measures:

\[
v_{i,j} = g(\tilde{s}_{i,j}) = \text{features}\left( CWT^{(1)}, CWT^{(2)}, \ldots, CWT^{(C)} \right)
\]

\subsubsection{Locally Linear Embedding (LLE):}

Locally Linear Embedding (LLE) is a technique that preserves the local linear structure of the data. For our vectorized normalized segments \( \mathbf{\tilde{s}}_{i,j} = \text{vec}(\tilde{s}_{i,j}) \in \mathbb{R}^{\tau C} \), LLE operates by reconstructing each segment from its nearest neighbors.

The steps are as follows:

\begin{enumerate}
    \item Find the set of \( K \) nearest neighbors \( \mathcal{N}_{i,j} \) for each segment \( \mathbf{\tilde{s}}_{i,j} \).
    \item Compute weights \( W_{i,j,k} \) that minimize the reconstruction error:
    \[
    \min_{W_{i,j}} \left\| \mathbf{\tilde{s}}_{i,j} - \sum_{k \in \mathcal{N}_{i,j}} W_{i,j,k} \mathbf{\tilde{s}}_{k} \right\|^2, \quad \text{subject to } \sum_{k \in \mathcal{N}_{i,j}} W_{i,j,k} = 1
    \]
    \item Compute the embeddings \( v_{i,j} \in \mathbb{R}^d \) by minimizing:
    \[
    \min_{v_{i,j}} \sum_{i,j} \left\| v_{i,j} - \sum_{k \in \mathcal{N}_{i,j}} W_{i,j,k} v_k \right\|^2
    \]
\end{enumerate}

This process results in embeddings that preserve local neighborhood structures of the original data.

\subsubsection{UMAP:}

Uniform Manifold Approximation and Projection (UMAP) is a dimensionality reduction technique that maps high-dimensional data into a lower-dimensional space while preserving both local and global structures. For the vectorized segments \( \mathbf{\tilde{s}}_{i,j} \), UMAP operates as follows:

\begin{enumerate}
    \item Compute the fuzzy simplicial set representation of the high-dimensional data based on a distance metric \( d(\mathbf{\tilde{s}}_{i,j}, \mathbf{\tilde{s}}_{k}) \).
    \item Optimize the low-dimensional embeddings \( v_{i,j} \in \mathbb{R}^d \) by minimizing the cross-entropy between the fuzzy simplicial sets of the high-dimensional and low-dimensional representations.
\end{enumerate}

The embedding function \( g \) is defined implicitly through this optimization:

\[
v_{i,j} = g(\mathbf{\tilde{s}}_{i,j}), \quad g: \mathbb{R}^{\tau C} \to \mathbb{R}^d
\]

\subsubsection{Graph Embedding:}

Graph Embedding learns low-dimensional representations of graphs by capturing their structural properties. For time series data, we construct a Visibility Graph (VG) from each segment \( \tilde{s}_{i,j} \).

In a Natural Visibility Graph (NVG), an edge between nodes \( n_i \) and \( n_j \) exists if:

\[
x(t_k) < x(t_i) + \frac{(x(t_j) - x(t_i))}{t_j - t_i} (t_k - t_i), \quad \forall t_k \in (t_i, t_j)
\]

The weight of the edge is:
\(
w_{ij} = \left| \frac{x(t_j) - x(t_i)}{t_j - t_i} \right|
\)

From the constructed graph \( G_{i,j} = (N_{i,j}, E_{i,j}) \), we extract features such as degree distributions, clustering coefficients, or apply graph embedding techniques like node2vec to obtain the embedding \( v_{i,j} \).

\subsubsection{Persistent Homology:}

Persistent Homology captures topological features by analyzing the birth and death of homological features across different scales. For each segment \( \tilde{s}_{i,j} \), we combine properties from Visibility Graphs and persistence diagrams.

The Horizontal Visibility Graph (HVG) condition is:

\[
x(t_i), x(t_j) > x(t_k), \quad \forall t_k \in (t_i, t_j)
\]

We compute persistence diagrams \( D_{i,j} \) from sublevel filtrations of \( \tilde{s}_{i,j} \). Features extracted include: Bottleneck distance to a reference diagram; \( p \)-Wasserstein distances; Betti curves: \(B_{i,j}(x) = \sum_{(b_k, d_k) \in D_{i,j}} \delta_{[b_k, d_k]}(x)\); Persistence entropy; Norms of the persistence landscape.

These features are combined with those from the visibility graphs to form the embedding \( v_{i,j} \).

\subsubsection{Autoencoder:}

Autoencoders learn compressed representations of data through unsupervised learning. For each normalized segment \( \tilde{s}_{i,j} \), the autoencoder consists of:

\begin{itemize}
    \item \textit{Encoder}:
    \(
    h_{i,j} = f_{\textnormal{encode}}(\tilde{s}_{i,j})
    \)
    \item \textit{Decoder}:
    \(
    \hat{\tilde{s}}_{i,j} = f_{\textnormal{decode}}(h_{i,j})
    \)
\end{itemize}

The embedding \( v_{i,j} \) is the encoded representation \( h_{i,j} \). The autoencoder is trained to minimize the reconstruction loss:

\[
\min_{f_{\textnormal{encode}}, f_{\textnormal{decode}}} \sum_{i,j} \left\| \tilde{s}_{i,j} - \hat{\tilde{s}}_{i,j} \right\|^2
\]

\subsubsection{Contrastive Learning CNN Embedding (CL-CNN):}

Each normalized segment \( \tilde{s}_{i,j} \) is transformed into an embedding vector \( v_{i,j} \) using a one-dimensional Convolutional Neural Network (1D-CNN). The CNN applies convolutional filters across the time dimension to extract temporal features.

The embedding process is defined as:

\[
v_{i,j} = \textnormal{CNN}(\tilde{s}_{i,j}), \quad \textnormal{CNN}: \mathbb{R}^{\tau \times C} \to \mathbb{R}^d
\]

where \( \textnormal{CNN} \) includes convolutional layers, activation functions, and pooling layers designed to capture hierarchical patterns in the data.

\subsubsection{Contrastive Learning RNN Embedding (CL-RNN):}

Each normalized segment \( \tilde{s}_{i,j} \) is processed using a Recurrent Neural Network (RNN) to capture temporal dependencies. The RNN updates its hidden state \( h_{i,j,k} \) at each time step \( k \):

\[
h_{i,j,k} = f_{\textnormal{RNN}}(s_{i,j,k}, h_{i,j,k-1}), \quad s_{i,j,k} \in \mathbb{R}^C, \quad h_{i,j,k} \in \mathbb{R}^h
\]

with \( h_{i,j,0} \) initialized appropriately. The final hidden state after processing the entire segment serves as the embedding:
\(
v_{i,j} = h_{i,j,\tau}
\).
This embedding captures sequential information from the entire window \( \tilde{s}_{i,j} \). In this work, an LSTM-based backbone was used as a recurrent neural network.

\smallskip
\textbf{Note}: To obtain \emph{unsupervised embeddings} using CNN and RNN-based models, we implement the \emph{nearest neighbor contrastive learning (NNCLR)} approach introduced by \cite{dwibedi2021little}, adapted for time series data. Therefore, we use the abbreviations \emph{CL-CNN} and \emph{CL-RNN} to refer to these embedding methods.


\subsection{Classification Algorithms}

To evaluate the effectiveness of the various embedding methods in capturing useful representations, we employ a range of widely used classification algorithms, as implemented in the Scikit-Learn library, introduced by \cite{pedregosa2011scikit}. These algorithms were chosen to represent different approaches to classification, allowing us to assess how well the embeddings perform across various learning paradigms. The classification algorithms used in this study are:

\begin{enumerate}
    \item \textit{Logistic Regression (LR)}: A linear model that estimates the probability of an instance belonging to a particular class.
    \item \textit{Decision Trees (DT)}: A non-parametric method that creates a model that predicts the target variable by learning simple decision rules inferred from the data features.
    \item \textit{Random Forest (RF)}: An ensemble learning method that operates by constructing multiple decision trees during training and outputting the class that is the mode of the classes of the individual trees.
    \item \textit{K-Nearest Neighbors (KNN)}: A non-parametric method that classifies a data point based on how its neighbors are classified.
    \item \textit{XGBoost (XGB)}: An optimized distributed gradient boosting library designed to be highly efficient, flexible, and portable.
    \item \textit{Support Vector Machines (SVM)}: A method that finds a hyperplane in an N-dimensional space that distinctly classifies the data points.
    \item \textit{Naive Bayes (NB)}: A probabilistic classifier based on applying Bayes' theorem with strong (naive) independence assumptions between the features.
    \item \textit{Multi-Layer Perceptron (MLP)}: A class of feedforward artificial neural networks that consist of at least three layers of nodes: an input layer, a hidden layer, and an output layer. An MLP, also known as a fully connected or dense neural network, usually forms the last few layers of a classification neural network (a.k.a., classification head), whereas previous layers act as complex feature extractors or feature learners. Using an MLP to classify an embedding essentially simulates this behavior.
\end{enumerate}

Each of these classification algorithms was applied to the embedded representations of the time series data produced by the various embedding methods. We used standard implementations of these algorithms in their respective libraries. To ensure that the best results per dataset and embedding method are considered for comparison, we used the Optuna library in Python, introduced by \cite{akiba2019optuna}, to tune the most important parameters of the classification methods.

As shown, the results indicate the average and standard deviation as a result of running the experiments for each time series embedding method and relative dataset.

\section{Results}
\label{sec:results}

Our experimental evaluation encompasses ten distinct time series embedding methods tested across eleven diverse datasets using various classification algorithms. The classification accuracies are presented in Table~\ref{table:embedding_results}, with averaged performance across all classifiers for each embedding method and dataset combination. The table also shows the average rank of each method. The rank was computed by our experimental evaluation, which encompasses ten distinct time series embedding methods tested across eleven diverse datasets using various classification algorithms. The classification accuracies are presented in Table~\ref{table:embedding_results}, with averaged performance across all classifiers for each embedding method and dataset combination. The table also shows the average rank of each method. The rank was computed by first ranking the embedding methods within each dataset based on their classification accuracy, where rank 1 corresponds to the highest accuracy and rank 10 to the lowest. For each dataset, ties in accuracy received the same rank, and the subsequent rank was adjusted accordingly (e.g., if two methods tied for rank 1, the next best method would receive rank 3). The average rank for each embedding method was then calculated by taking the arithmetic mean of its ranks across all eleven datasets. This ranking approach provides a robust measure of overall performance that accounts for the relative effectiveness of each embedding method across diverse signal types and application domains, with lower average ranks indicating better overall performance.

\begin{table*}
\captionsetup[table]{skip=5pt} 
\centering
\caption{\small Comparison of classification accuracies based on the embedding method. Each value shows the average accuracy and standard deviation that the embedding method yielded for all classification algorithms on the corresponding dataset.}
\begin{adjustbox}{width=\textwidth}
\renewcommand{\arraystretch}{1.3} 
\setlength{\tabcolsep}{12pt} 
\begin{tabular}{l|c|c|c|c|c|c|c|c|c|c}
\hline
\textbf{Dataset} & \textbf{PCA} & \textbf{Wavelet} & \textbf{FFT} & \textbf{LLE} & \textbf{UMAP} & \textbf{Graph} & \textbf{TDA}  & \textbf{AE} & \textbf{C-CNN} & \textbf{C-RNN}   \\ \specialrule{1.2pt}{0pt}{0pt}
Sleep & 0.685 & \textbf{0.715} & 0.698 & 0.645 & 0.665 & 0.673 & 0.638 & 0.625 & 0.667 & 0.623  \\ 
ElectricDevices & \textbf{0.572} & 0.563 & 0.568 & 0.542 & 0.555 & 0.548 & 0.535 & 0.525 & 0.542 & 0.515 \\ 
MelbournePedestrian & 0.662 & 0.655 & \textbf{0.685} & 0.625 & 0.648 & 0.652 & 0.592 & 0.585 & 0.602 & 0.562 \\ 
Racketsport & 0.708 & \textbf{0.728} & 0.715 & 0.675 & 0.685 & 0.690 & 0.622 & 0.638 & 0.672 & 0.602 \\ 
SharePriceIncrease & \textbf{0.695} & 0.679 & 0.689 & 0.621 & 0.636 & 0.679 & 0.683 & 0.643 & 0.661 & 0.655  \\
SelfRegulationSCP1 & 0.745 & \textbf{0.782} & 0.762 & 0.705 & 0.728 & 0.698 & 0.685 & 0.675 & 0.715 & 0.658 \\
UniMib & 0.754 & \textbf{0.777} & 0.709 & 0.761 & 0.650 & 0.650 & 0.633 & 0.551 & 0.664 & 0.411 \\
EMGGestures & 0.615 & \textbf{0.668} & 0.642 & 0.592 & 0.605 & 0.622 & 0.585 & 0.562 & 0.635 & 0.535 \\
Mill & 0.899 & 0.826 & \textbf{0.909} & 0.809 & 0.851 & 0.812 & 0.766 & 0.776 & 0.834 & 0.715 \\
ECG5000 & 0.923 & 0.925 & \textbf{0.927} & 0.911 & 0.901 & 0.920 & 0.681 & 0.741 & 0.902 & 0.892 \\\hline
Avg Rank & 2.6 & 2.2 & \textbf{1.9} & 6.2 & 5.5 & 5.0 & 7.9 & 8.7 & 5.5 & 9.5 \\ \hline

\end{tabular}
\end{adjustbox}
\label{table:embedding_results}
\end{table*}


\subsection{Overall Performance}
The experimental results demonstrate that embedding method performance varies significantly across different datasets and classification algorithms. PCA consistently delivers strong performance (average rank 2.6), while Wavelet Transform (average rank 2.2) and FFT (average rank 1.9) show the best overall performance across all datasets. Advanced methods like TDA and Graph Embedding show competitive performance on specific datasets but exhibit higher variability. Among the deep learning approaches, C-CNN (average rank 5.5) outperforms C-RNN (average rank 9.5) across most datasets, suggesting that convolutional architectures may be more effective at capturing relevant temporal patterns for classification tasks when applied directly to raw data. The overall low ranking of deep-learning-based methods can be justified by the fact that self-supervised learning methods, such as the NNCLR \cite{dwibedi2021little} method used in this study, require large amounts of data and extensive hyperparameter tuning to be trained effectively.

\subsection{UMAP projection}
\begin{figure}
\vspace{-0.5cm}
    \centering
    \includegraphics[width=0.4\linewidth]{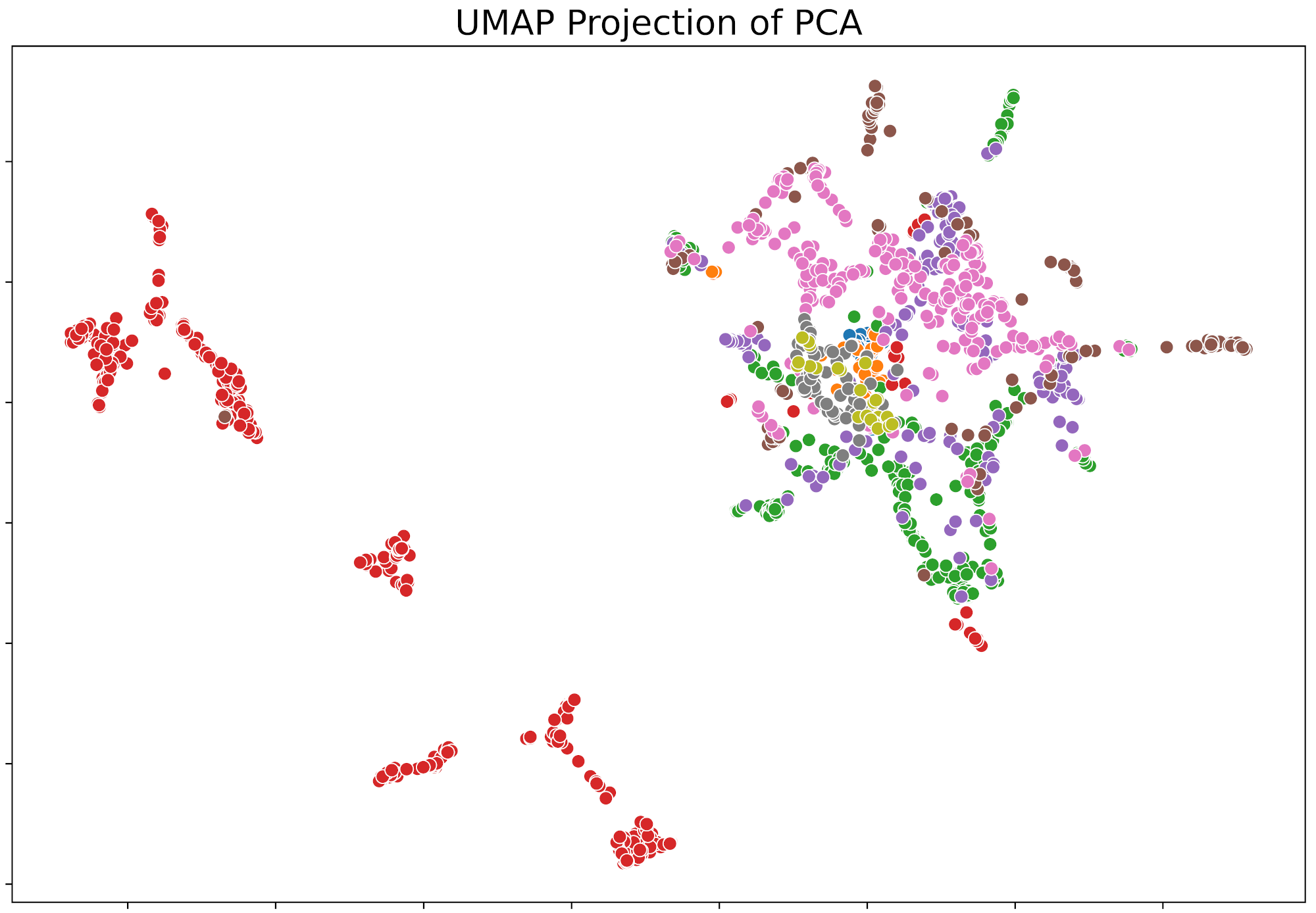} \hfill
    \includegraphics[width=0.4\linewidth]{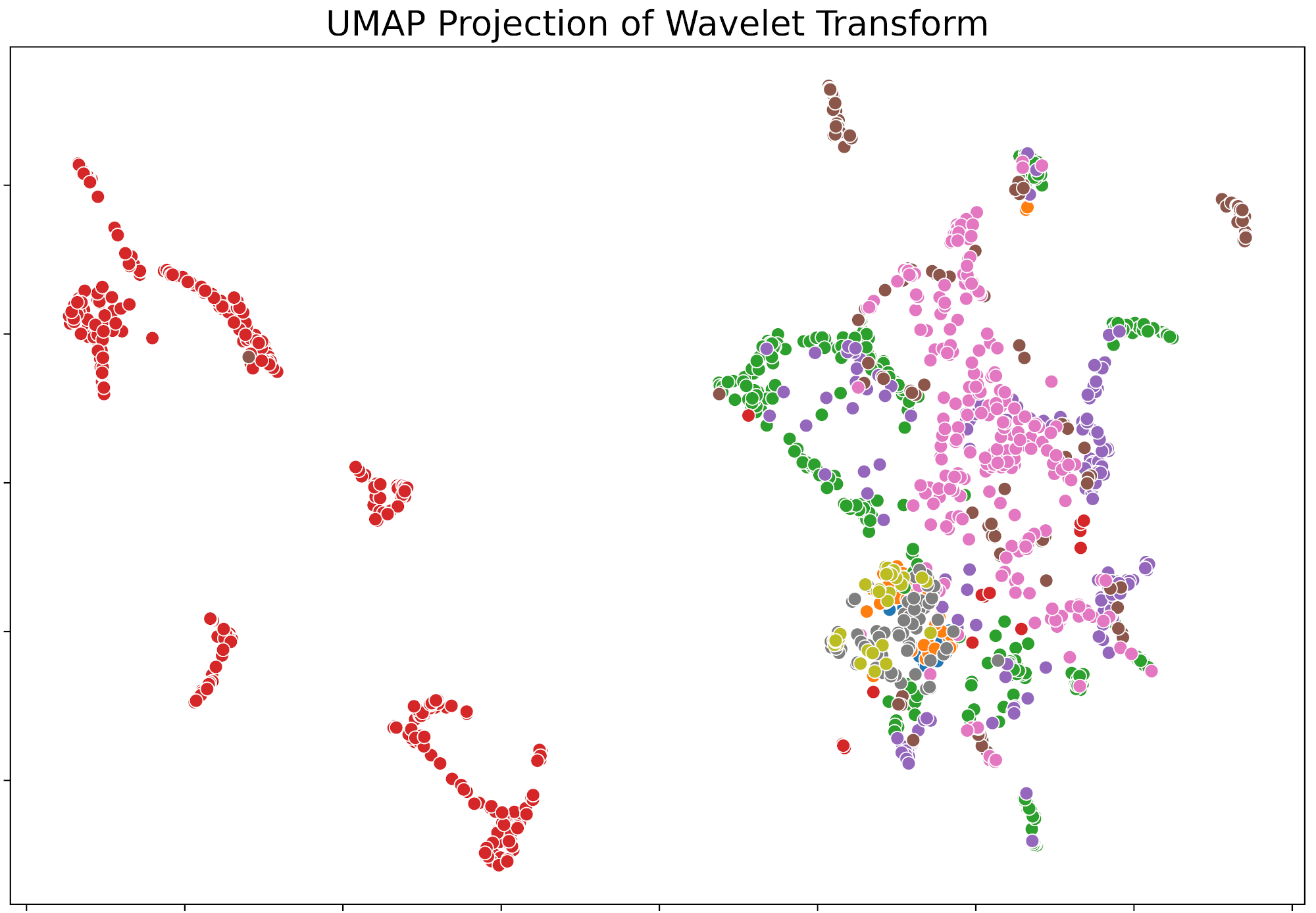} \\
    \includegraphics[width=0.4\linewidth]{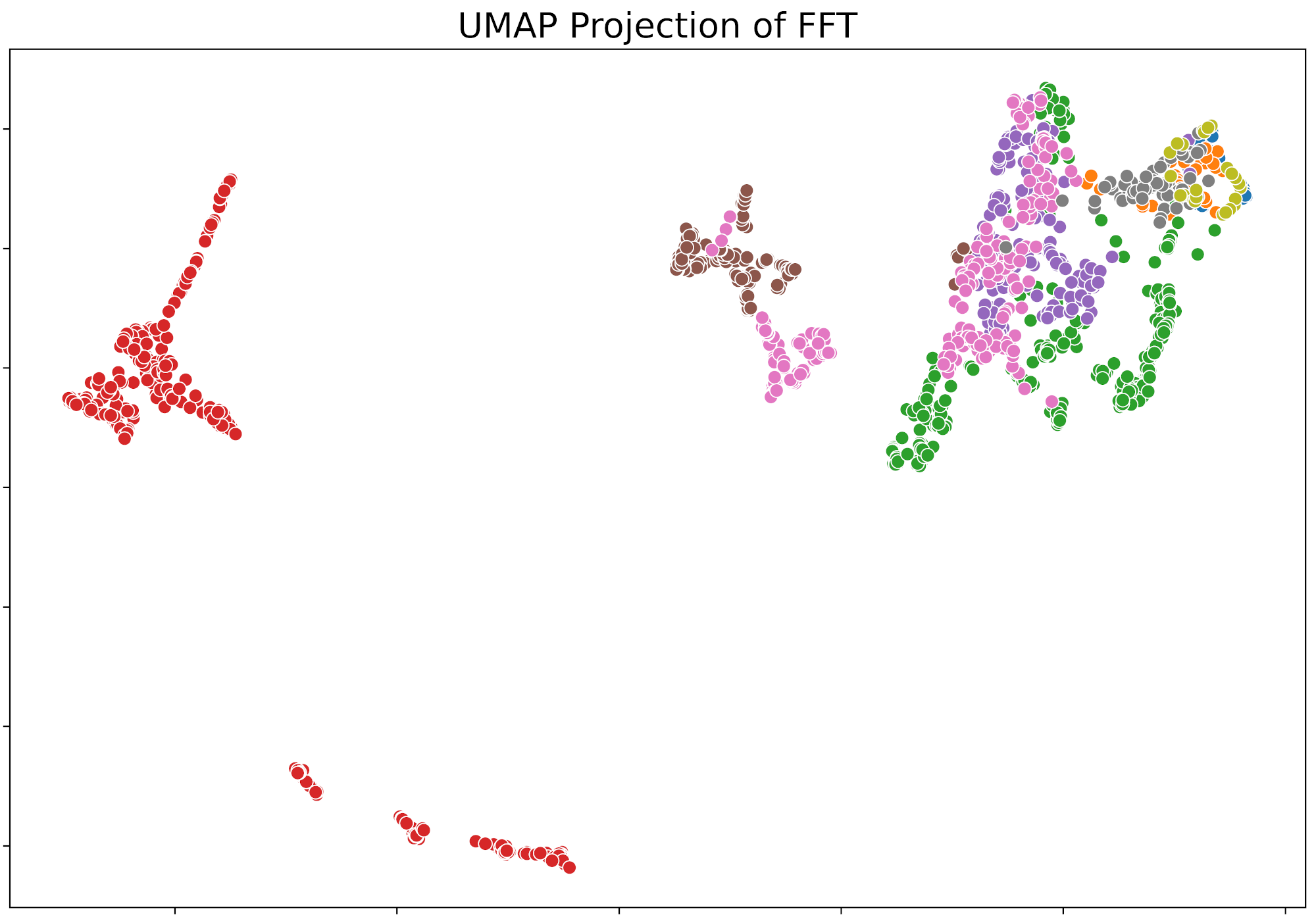} \hfill
    \includegraphics[width=0.4\linewidth]{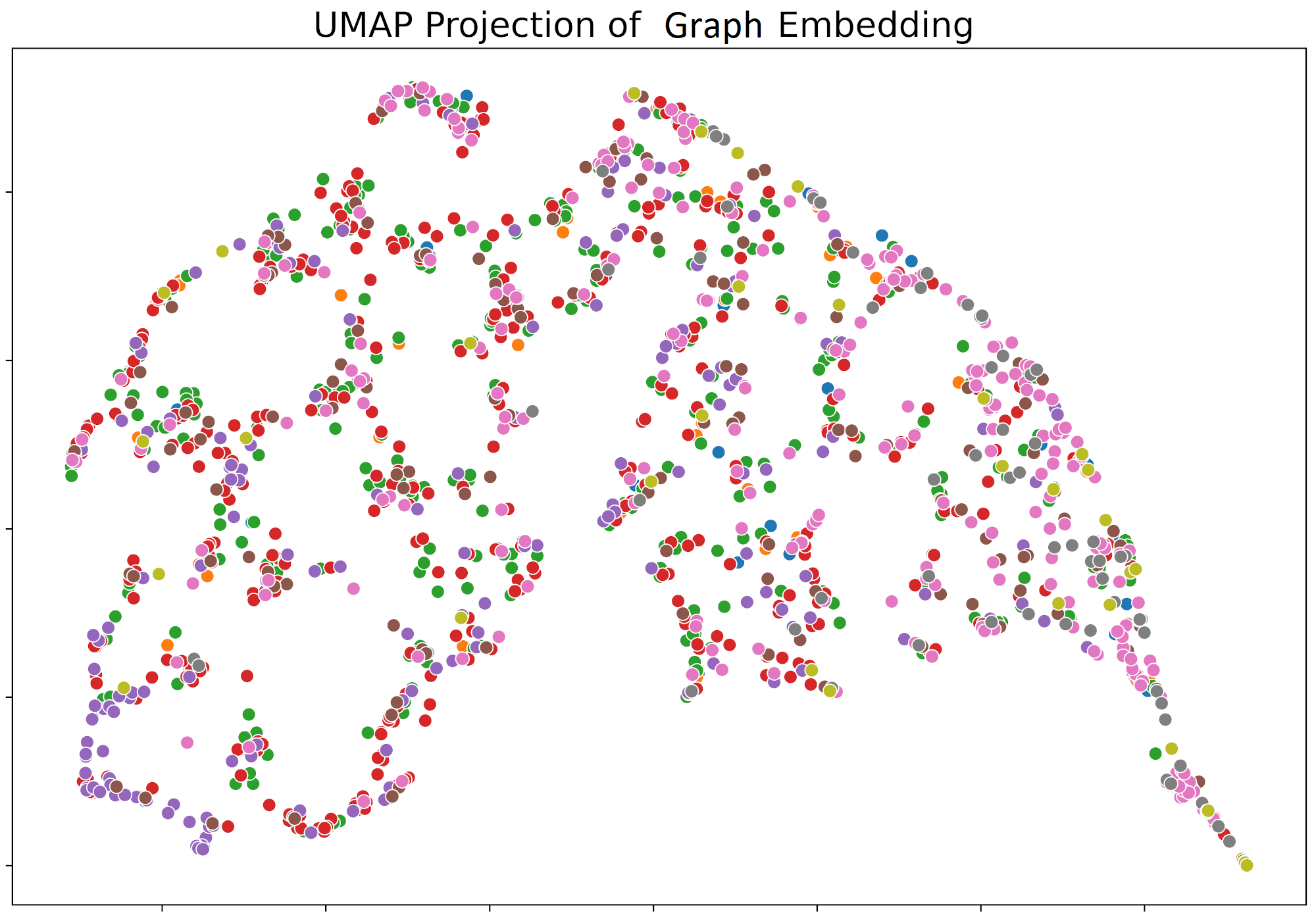}\\
    \includegraphics[width=0.4\linewidth]{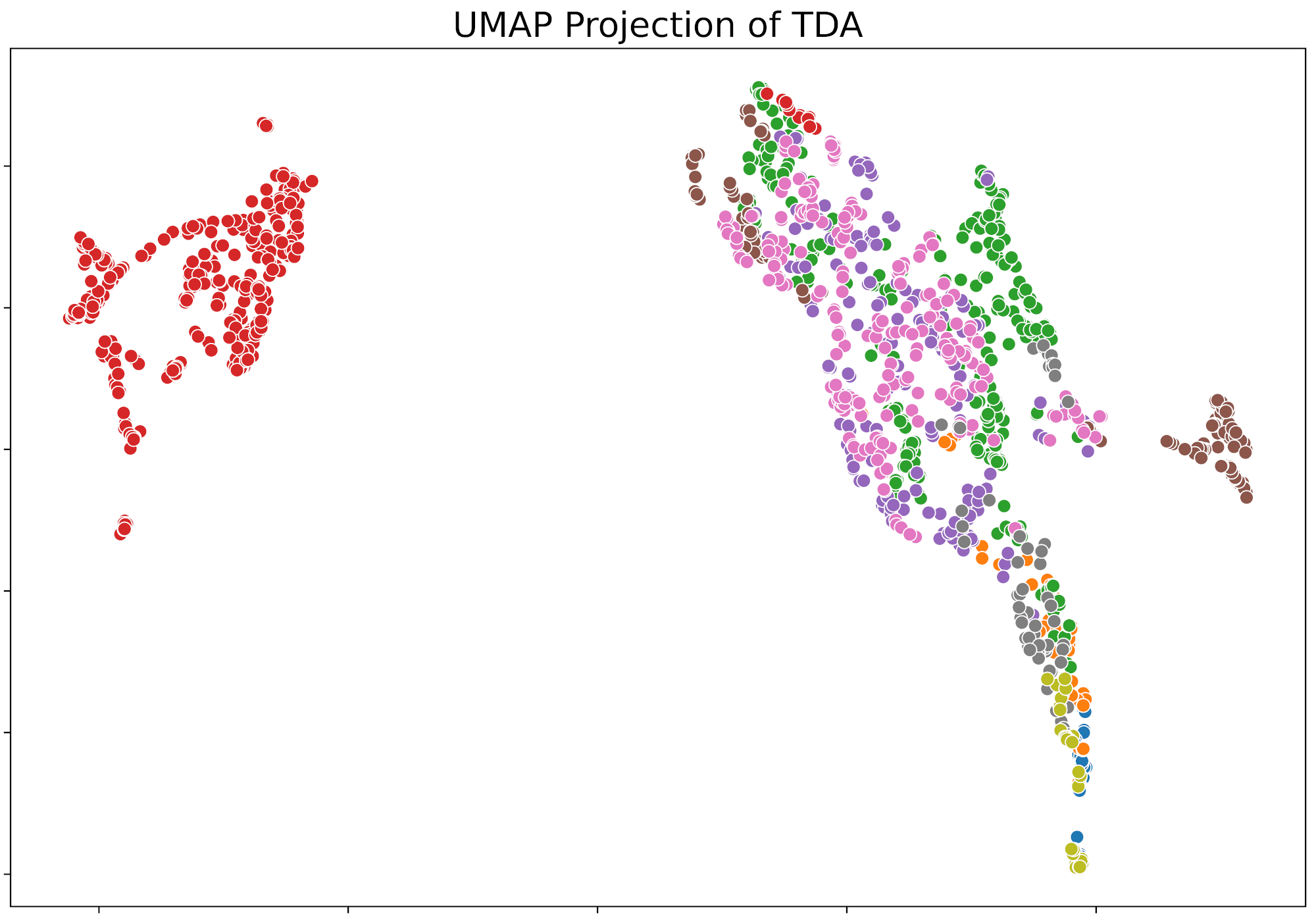} \hfill
    \includegraphics[width=0.4\linewidth]{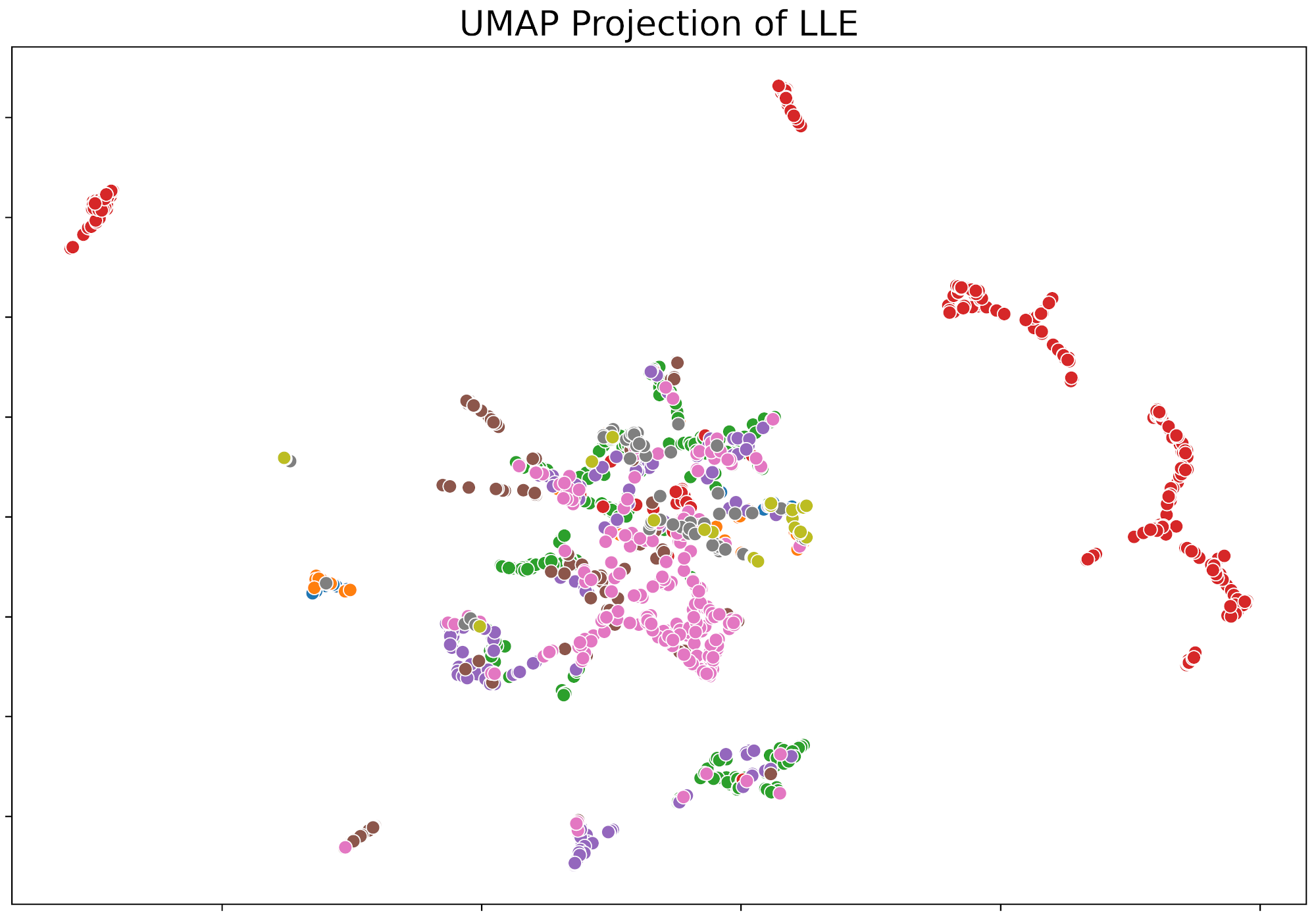}\\
    \includegraphics[width=0.4\linewidth]{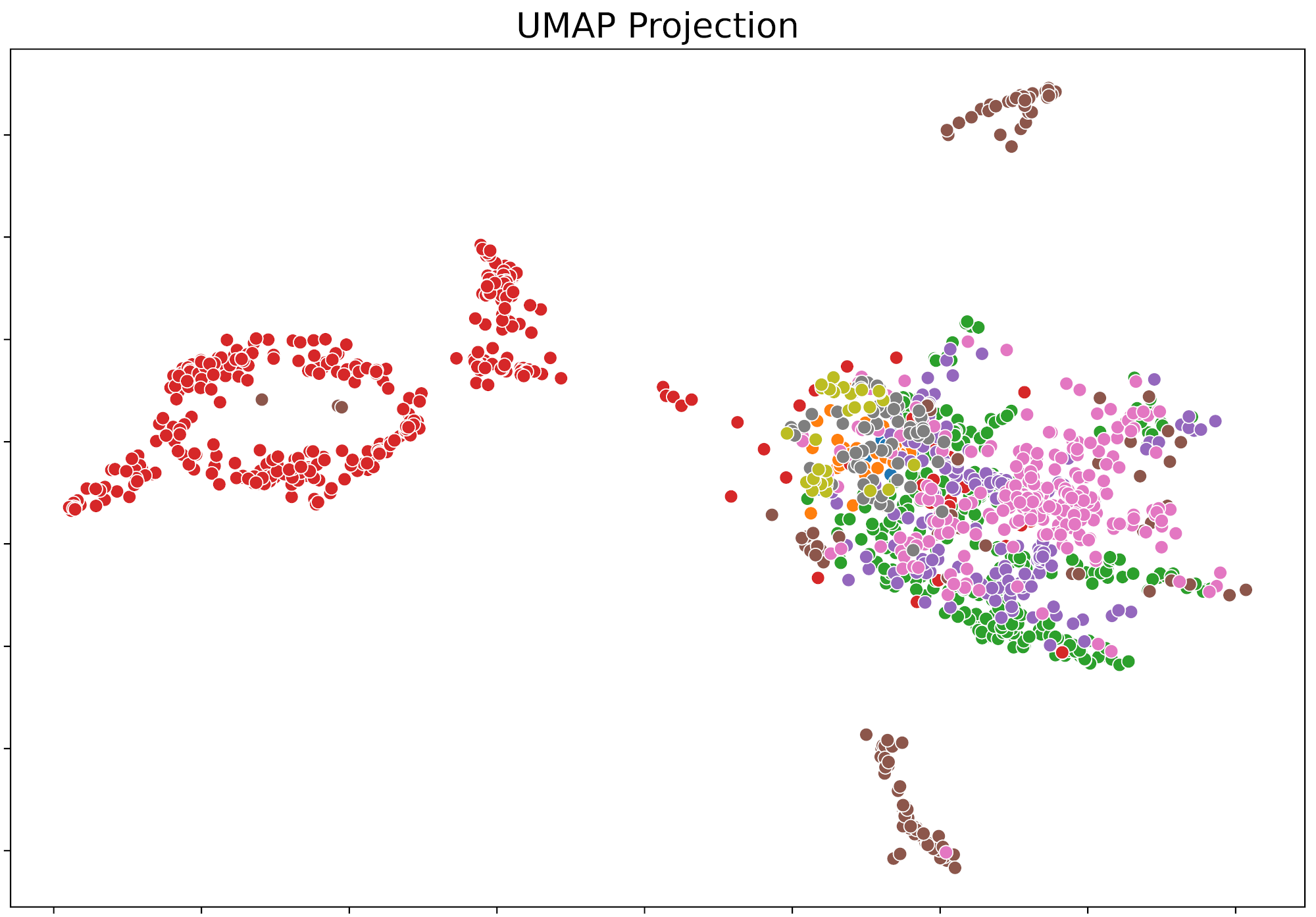} \hfill
    \includegraphics[width=0.4\linewidth]{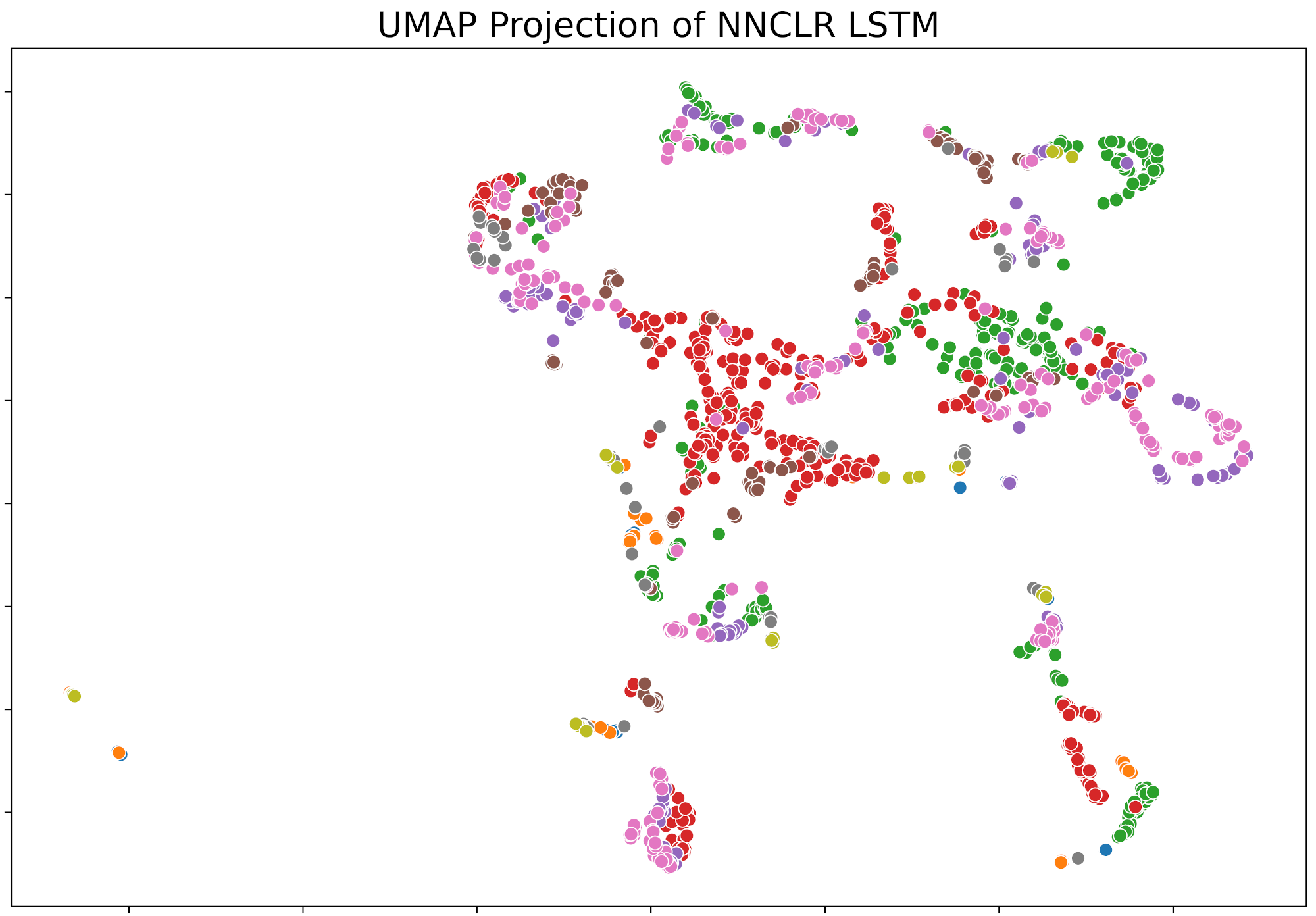}\\
    \includegraphics[width=0.4\linewidth]{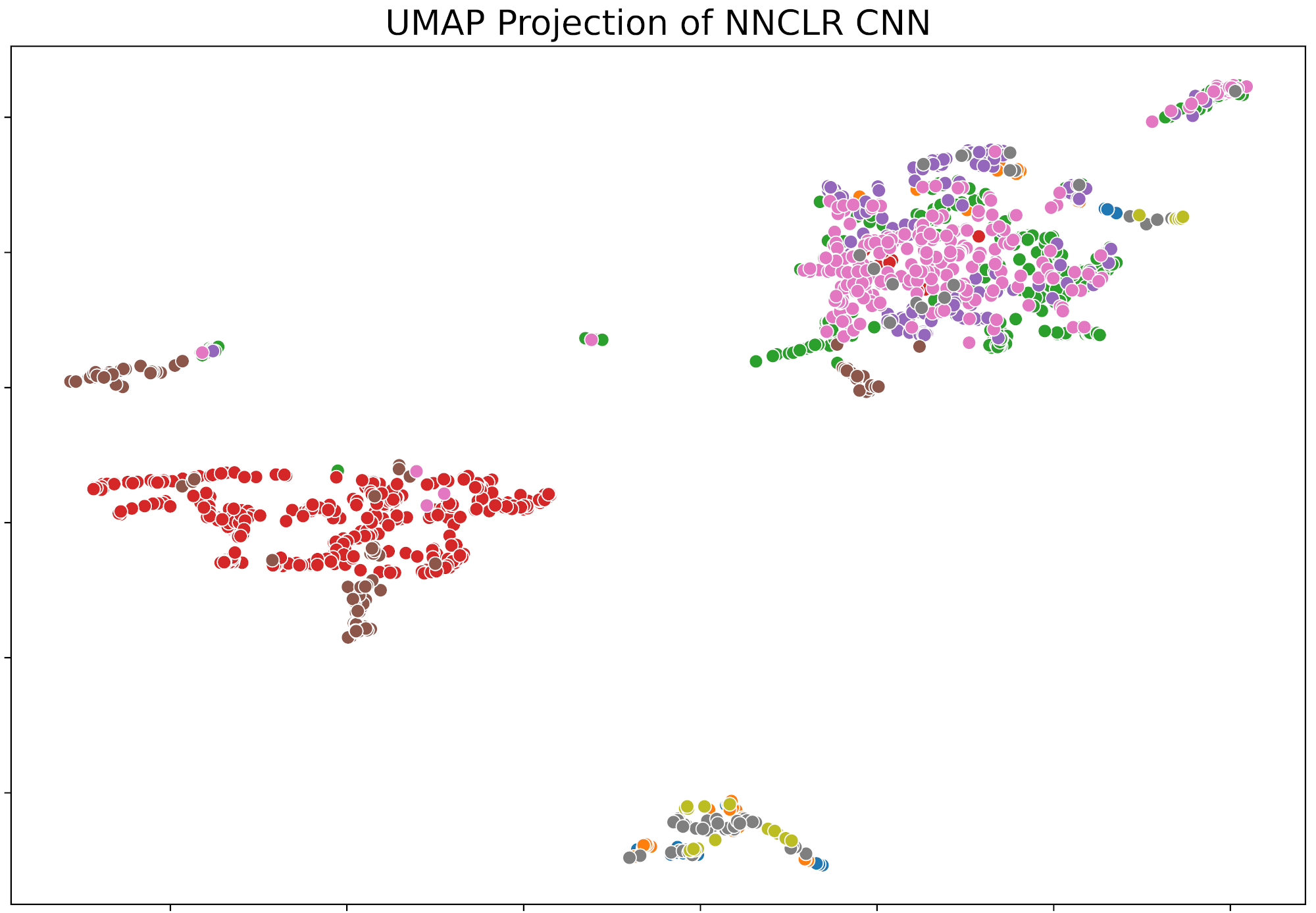} \hfill
    \includegraphics[width=0.4\linewidth]{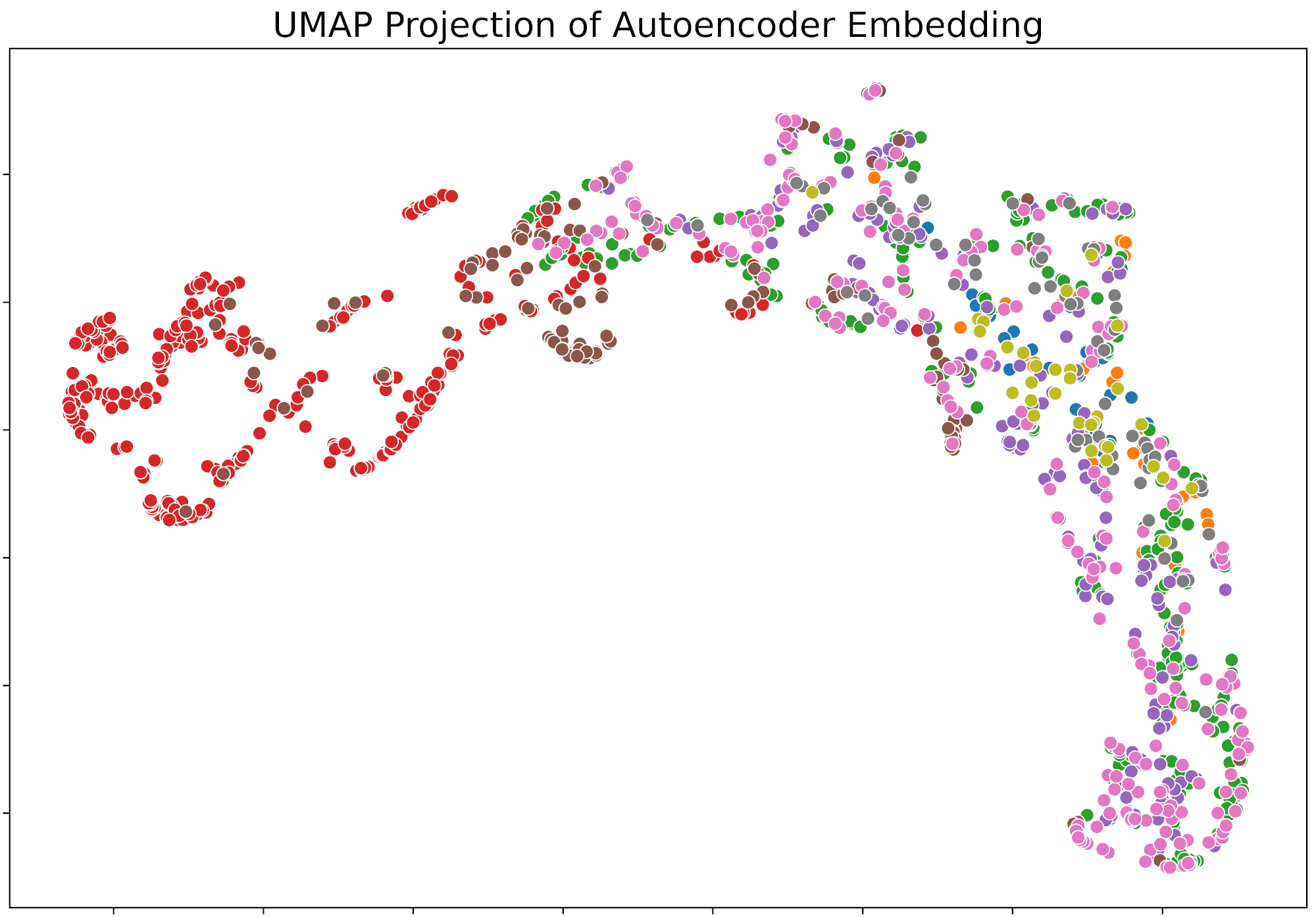}
  \caption{UMAP projections of different embedding methods on the UniMiB dataset.}
  \label{fig:umap_projections}
\end{figure}

For an initial visual qualitative overview of the embeddings produced by each method, we have plotted their UMAP projections on the UniMiB SHAR dataset in Figure~\ref{fig:umap_projections}. The data points are color-coded by their class label. Better visual separation of the data points from different classes likely means that the downstream classifier will have an easier time correctly classifying the data. However, it should be noted that the separability also depends on the ability of the UMAP projection to preserve the embedding properties when projecting from d-dimensions to two dimensions.

\subsection{Dataset Complexity Analysis}
The effectiveness of embedding methods shows a strong correlation with dataset characteristics, particularly dimensionality and sequence length. For datasets with high dimensionality (more than 5 channels) such as EMGGestures and RacketSports, Wavelet Transform consistently outperforms other methods, achieving accuracies of 66.8\% and 72.8\% respectively. This suggests that Wavelet's multi-resolution capabilities are particularly beneficial for capturing complex relationships across multiple channels. Conversely, for univariate time series such as ECG5000 and ElectricDevices, FFT and PCA demonstrate superior performance, with accuracies reaching 92.7\% and 57.2\% respectively.

\subsection{Domain-Specific Performance}

\subsubsection{Bioelectrical Signals}
In EEG datasets (Sleep, SelfRegulationSCP1), Wavelet Transform demonstrates optimal performance, achieving accuracies of 71.5\% and 78.2\% respectively. LLE also performs strongly on SelfRegulationSCP1 (70.5\%), while PCA (68.5\%) and FFT (69.8\%) show competitive performance on Sleep data. For ECG data (ECG5000), FFT achieves the highest accuracy (92.7\%), closely followed by Wavelet Transform (92.5\%) and PCA (92.3\%). These results suggest that frequency-domain representations are particularly effective for capturing the quasi-periodic components characteristic of bioelectrical signals.

\subsubsection{Biomechanical and Motion Signals}
For biomechanical signals (UniMiB-SHAR, RacketSports, EMGGestures), Wavelet Transform achieves the highest accuracies (77.7\% for UniMiB-SHAR, 72.8\% for RacketSports, 66.8\% for EMGGestures respectively). The multi-resolution capability of wavelets appears particularly beneficial for analyzing the hierarchical temporal patterns in human movement data. UMAP also performs strongly on these datasets, indicating that manifold learning approaches can effectively capture the underlying nonlinear dynamics of biomechanical systems. Graph Embedding (62.2\% for EMGGestures) demonstrates moderate effectiveness in representing the structural relationships in muscle activation patterns, while deep learning approaches struggle to match the performance of classical methods in this domain.

\subsubsection{Electrical and Mechanical System Signals}
For electrical system datasets (ElectricDevices) and mechanical system datasets (Mill), FFT and PCA demonstrate superior performance, with FFT achieving the highest accuracy on Mill data (90.9\%). This confirms the efficacy of frequency-domain analysis for systems with characteristic spectral signatures and harmonic components. PCA's strong performance (89.9\% for Mill) suggests that linear subspace projection methods can effectively capture the dominant modes of variation in mechanical system signals. Graph Embedding also shows competitive performance on electrical system data (54.8\% for ElectricDevices), indicating that structural approaches can effectively represent the temporal state transitions in these systems. The relative underperformance of topological methods (76.6\% for Mill) suggests limitations in capturing the specific periodicity and harmonic structures of mechanical systems through persistence features alone.

\subsubsection{Economic and Environmental Signals}
For economic signals (SharePriceIncrease) and environmental signals (MelbournePedestrian), PCA achieves the highest accuracy for financial data (69.5\%), while FFT performs best for pedestrian traffic (68.5\%). The effectiveness of linear methods for economic time series suggests that dimensional reduction techniques can effectively isolate the latent factors driving financial markets. Graph Embedding shows comparable performance on financial data (67.9\%), potentially capturing the complex state transitions and regime shifts characteristic of economic systems. For traffic flow signals, which exhibit both periodic and stochastic components, the frequency-domain representation offered by FFT appears particularly effective at isolating seasonal and daily patterns. The overall modest performance across methods for these domains highlights the inherent challenge in modeling systems with both deterministic and stochastic components.

\subsection{Computational Efficiency Analysis}
While the representational power of embedding methods is a primary consideration, computational efficiency is also a critical factor for practical signal processing applications. Our analysis reveals significant variations in computational requirements across methods. Classical methods like PCA and FFT are highly efficient, requiring minimal computational resources even for long sequences. In contrast, manifold learning methods (LLE, UMAP) and deep learning approaches incur substantially higher computational costs, with self-supervised deep learning methods requiring up to 1,600$\times$ longer training times than PCA for the electricDevices dataset. Table~\ref{tab:timing_results} demonstrates the computational trade-offs across embedding methods on the ElectricDevices dataset, revealing that while deep learning models achieve excellent inference speeds (0.14-0.15s), they require substantial training investments of approximately 11-12 minutes each. Graph Embedding and TDA methods show moderate to high computational requirements but process all data identically without traditional training/inference distinctions.

\begin{table}
\centering
\caption{Training and Inference Times of different embedding methods on the ElectricDevices dataset.}
\begin{adjustbox}{width=0.45\columnwidth}
\centering
\setlength{\tabcolsep}{4pt} 
\begin{tabular}{@{}lccc@{}}
\toprule
\textbf{Embedding Method} & \textbf{Training (s)} & \textbf{Inference (s)} \\
\midrule
PCA & 0.443 & 0.064  \\
Wavelet Transform & 0.010 & 0.005   \\
FFT & 0.103 & 0.090   \\
LLE & 23.377 & 1.604   \\
UMAP & 21.719 & 15.000  \\
Graph & 364.340 & 309.867  \\
TDA & 204.619 & 176.771  \\
Autoencoder & 38.609 & 1.181  \\
C-CNN (200 epochs) & 680.504 & 0.139   \\
C-RNN (200 epochs) & 703.063 & 0.150   \\
\bottomrule
\end{tabular}
\label{tab:timing_results}
\end{adjustbox}
\end{table}

\subsection{Classification Algorithm Impact}
The effectiveness of embedding methods as shown in table \ref{table:classification_results} varies significantly depending on the downstream classification algorithm, highlighting the importance of considering the entire signal processing pipeline. Tree-based methods (Random Forest and XGBoost) consistently outperform other classifiers across most embedding methods, with Random Forest achieving particularly strong results on frequency-domain embeddings (FFT) with an average rank of 1.9. SVM also demonstrates competitive performance, particularly with manifold learning embeddings like LLE and UMAP. These results highlight the importance of considering the entire pipeline when selecting embedding methods for time series classification tasks.

\begin{table}[ht]
\centering
\caption{\small Comparison of classification accuracies based on the classification algorithm. Each value shows the average accuracy and standard deviation that the classification algorithm yielded for all embedding methods on the corresponding dataset.}
\begin{adjustbox}{width=\textwidth}
\renewcommand{\arraystretch}{1.3} 
\setlength{\tabcolsep}{12pt} 
\begin{tabular}{l|c|c|c|c|c|c|c|c}
\hline
\textbf{Dataset} & \textbf{LR} & \textbf{DT} & \textbf{RF} & \textbf{KNN} & \textbf{XGB} & \textbf{SVM} & \textbf{NB}  & \textbf{MLP}  \\ \specialrule{1.2pt}{0pt}{0pt}
Sleep & 0.665 & 0.651 & \textbf{0.697} & 0.677 & 0.691 & 0.678 & 0.556 & 0.654 \\    
ElectricDevices  & 0.553 & 0.539 & 0.565 & 0.558 & \textbf{0.581} & 0.560 & 0.498 & 0.552  \\ 
MelbournePedestrian & 0.572 & 0.558 & 0.618 & 0.602 & \textbf{0.632} & 0.621 & 0.515 & 0.592 \\ 
Racketsport & 0.655 & 0.632 & 0.710 & 0.683 & 0.718 & \textbf{0.723} & 0.588 & 0.631  \\ 
SharePriceIncrease  & 0.694 & 0.647 & 0.684 & 0.671 & 0.661 & \textbf{0.692} & 0.594 & 0.670  \\
SelfRegulationSCP1  & 0.715 & 0.702 & 0.749 & 0.723 & \textbf{0.752} & 0.732 & 0.655 & 0.695 \\
UniMib & 0.640 & 0.620 & 0.701 & 0.693 & 0.708 & \textbf{0.710} & 0.549 & 0.628 \\
EMGGestures  & 0.595 & 0.582 & \textbf{0.665} & 0.618 & 0.658 & 0.625 & 0.538 & 0.610  \\
Mill & 0.711 & 0.893 & 0.925 & 0.907 & \textbf{0.929} & 0.797 & 0.596 & 0.803 \\
ECG5000 & 0.856 & 0.874 & \textbf{0.904} & 0.876 & 0.902 & 0.887 & 0.825 & 0.843\\ \hline
Avg Rank & 5.1 & 6.3 & 2.1 & 3.9 & \textbf{2.0} & 2.7 & 8.0 & 5.9  \\ \hline
\end{tabular}
\end{adjustbox}
\label{table:classification_results}
\end{table}

\section{Discussion}
\subsection{Impact of Signal Characteristics}
Our comprehensive analysis reveals that the effectiveness of embedding methods is fundamentally influenced by time series characteristics, particularly their dimensionality, sequence length, and application domain. For high-dimensional multivariate signals (or time series), Wavelet Transform consistently excels, likely due to its ability to capture both time and frequency information across multiple channels simultaneously. The multi-resolution nature of wavelets enables the representation of temporal patterns at different scales, which is particularly valuable for complex physiological signals and human activity data.

Conversely, for univariate or low-dimensional signals, simpler methods like PCA and FFT often achieve comparable or superior performance to more complex approaches. This suggests that for datasets with simpler structures, the additional complexity of advanced embedding methods may not translate to proportionate performance gains. The strong performance of FFT across multiple domains indicates that frequency-domain representations remain highly effective for a wide range of time series classification tasks, particularly those involving periodic or quasi-periodic signals.

The relationship between time series length and embedding method effectiveness also reveals important patterns. For longer time series (or signals), methods that can effectively compress information, such as PCA and Wavelet Transform, demonstrate advantages over methods that struggle with the curse of dimensionality. This pattern becomes particularly evident in datasets like SelfRegulationSCP1, where dimensionality reduction becomes essential for effective classification.

\subsection{Methodological Considerations}

The comparative analysis across embedding categories reveals distinct advantages and limitations that have important implications for method selection. Classical methods (PCA, FFT, Wavelet Transform) offer robust performance, computational efficiency, and interpretability, making them valuable baseline approaches for many applications. Their consistent performance across diverse datasets suggests that they should be considered as strong candidates before implementing more complex methods.

Machine learning-based methods (LLE, UMAP) excel at capturing non-linear relationships and complex manifold structures, but their performance varies significantly across datasets. These methods show particular promise for datasets with complex underlying geometries, such as human activity recognition, but may struggle with noisy or irregularly sampled time series. Their higher computational requirements also present challenges for large-scale applications.

Structural and topological methods (Graph Embedding, TDA) show mixed results, with performance varying substantially across signal types. Graph-based approaches demonstrate particular strengths for signals with distinct state transitions or regime shifts, suggesting potential applications in anomaly detection and change point analysis. However, their overall performance lags behind classical methods for standard classification tasks, indicating that structural features alone may not provide sufficient discriminative power for many signal types.

Deep learning-based methods show surprisingly modest performance, with even the best-performing approach (C-CNN) generally not matching classical techniques. This finding suggests that self-supervised representation learning, while theoretically promising, may face practical challenges in extracting discriminative features from time series without extensive architecture optimization and large training datasets. The substantial gap between C-CNN and C-RNN performance indicates that architectural choices significantly impact representation quality, with convolutional approaches better suited to capturing the relevant temporal patterns for many signal types when applied directly to raw time series. In practice, hybrid architectures, which include convolutional layers closer to the input and recurrent or Transformer layers deeper in the network, may be the best approach.

\subsection{Classification Algorithm Synergies}
The interaction between embedding methods and classification algorithms reveals important synergistic effects that significantly impact overall system performance. The consistently strong performance of tree-based methods across multiple embedding types suggests that these classifiers possess intrinsic advantages for time series classification, likely stemming from their ability to identify discriminative features from diverse representations and their robustness to irrelevant or noisy dimensions.

The particular synergy between manifold embeddings and SVM classification highlights the value of geometric preservation in the embedded space. By maintaining the topological structure of the original signal manifold, methods like LLE and UMAP provide representations that align well with SVM's margin-based optimization, resulting in superior classification boundaries. This geometric perspective offers valuable insights for signal classification system design, suggesting that preserving the intrinsic structure of the signal manifold may be more important than maximizing variance or minimizing reconstruction error.

These synergistic relationships highlight the importance of considering the entire signal processing pipeline when developing classification systems. Rather than evaluating embedding methods in isolation, optimal performance requires joint optimization of both representation and classification components, with particular attention to their mutual compatibility. This system-level perspective aligns with modern signal processing frameworks that emphasize end-to-end optimization rather than individual component excellence.

\subsection{Practical Guidelines for Signal Processing Applications}
Based on our comprehensive evaluation, we propose the following practical guidelines for selecting time series embedding methods in signal processing applications:

\begin{enumerate}
    \item \textbf{Prioritize classical methods for most applications}: For the majority of time series classification tasks, classical methods like Wavelet Transform, FFT, and PCA should be evaluated first due to their robust performance, computational efficiency, and interpretability. These methods provide strong baselines that more complex approaches must justify surpassing.
    
    \item \textbf{Select methods based on signal characteristics}: For signals with strong harmonic components or clear spectral signatures (e.g., ECG, mechanical vibrations), FFT typically provides optimal results. For non-stationary signals with transient features (e.g., EEG, speech), Wavelet Transform offers superior performance. For signals with complex nonlinear dynamics (e.g., human motion, fluid dynamics), consider manifold learning approaches despite their higher computational costs.
    
    \item \textbf{Match techniques to application domains}: For bioelectrical signal processing, wavelet-based representations consistently excel. For mechanical and electrical system analysis, frequency-domain representations (FFT) and principal component analysis (PCA) provide the most effective embeddings. For biomechanical signal analysis, consider combined approaches that capture both spectral and geometric properties.
    
    \item \textbf{Optimize the full processing pipeline}: Consider both embedding method and classification algorithm when designing signal processing systems. Tree-based methods (Random Forest, XGBoost) offer robust performance across most embedding approaches. For manifold embeddings, SVM typically provides superior classification. For wavelet embeddings, neural network classifiers show particular promise.
\end{enumerate}

These guidelines aim to assist signal processing practitioners in navigating the complex landscape of time series embedding methods, enabling more informed decisions based on specific signal characteristics and application requirements.

\section{Conclusion}
\label{sec:conclusion}

This comprehensive study evaluates various time series embedding methods across different datasets and classification tasks, revealing important insights into their relative strengths and limitations. Our analysis demonstrates that while embedding method performance varies significantly based on dataset characteristics and downstream tasks, classical methods like PCA and Fourier transforms consistently offer robustness and interpretability for datasets with prominent global patterns. In contrast, complex methods such as deep learning-based embeddings excel at capturing non-linear patterns in datasets with intricate structures, though at higher computational costs.

The selection between classical and complex embedding methods inherently involves trade-offs between simplicity, interpretability, computational efficiency, and pattern-capturing ability. Our findings emphasize the importance of adopting a tailored approach that carefully considers the specific characteristics of the data and intended analysis goals. The varying effectiveness of topological and graph-based methods across different applications suggests promising avenues for future development, particularly in handling complex, multi-dimensional time series data.

Through the provision of an open-source suite implementing these embedding methods, we aim to facilitate further advancements in time series analysis across various fields. Future research directions include developing hybrid and adaptive embedding methods, improving interpretability of complex techniques, and extending the evaluation to other domains, ultimately contributing to the broader understanding and application of these tools.






\bibliographystyle{unsrtnat}
\bibliography{refs}

\end{document}